\documentclass[times,twocolumn,final]{elsarticle}

\usepackage{medima}
\usepackage{framed,multirow}

\usepackage{amssymb}
\usepackage{latexsym}

\usepackage{url}

\usepackage{hyperref}

\usepackage{amsmath,amssymb,amsfonts}
\usepackage{algorithmic}
\usepackage{graphicx}
\usepackage{algorithm,algorithmic}
\usepackage{hyperref}
\usepackage{textcomp}
\usepackage{url}
\usepackage{threeparttable}
\usepackage{booktabs}
\usepackage{multirow}
\usepackage{float}

\journal{****}

\begin{document}

\verso{Given-name Surname \textit{et~al.}}

\begin{frontmatter}

\title{Dual Graph Attention-based Disentanglement Multiple Instance Learning for Brain Age Estimation}%

\author[1]{Fanzhe Yan\corref{cor1}}
\cortext[cor1]{Corresponding author.}
\ead{yanfanzhe@mail.ustc.edu.cn}
\author[1]{Gang Yang}
\author[1]{Yu Li}
\author[1]{Aiping Liu}
\author[1]{Xun Chen}

\address[1]{Department of Electronic Engineering and Information Science, University of Science and Technology
of China, Hefei 230027, People’s Republic of China}

\received{}
\finalform{}
\accepted{}
\availableonline{}
\communicated{}

\begin{abstract}
Deep learning techniques have demonstrated great potential for accurately estimating brain age by analyzing Magnetic Resonance Imaging (MRI) data from healthy individuals. However, current methods for brain age estimation often directly utilize whole input images, overlooking two important considerations: 1) the heterogeneous nature of brain aging, where different regions contribute to the aging process at different degrees, and 2) the existence of age-independent redundancies in brain structure. To overcome these limitations, we propose a Dual Graph Attention-based Disentanglement Multi-instance Learning (DGA-DMIL) framework for improving brain age estimation. Specifically, the 3D MRI data, treated as a bag of instances, is fed into a 2D convolutional neural network backbone, to capture the aging patterns in MRI. A dual graph attention aggregator is then proposed to learn the backbone features by exploiting the intra- and inter-instance relationships to provide the instance contribution scores explicitly. Furthermore, a disentanglement branch is introduced to separate age-related features from age-independent structural representations to ameliorate the interference of redundant information on age prediction. To verify the effectiveness of the proposed framework, we evaluate it on two datasets, UK Biobank and ADNI, containing a total of 35,388 healthy individuals. Our proposed model demonstrates exceptional accuracy in estimating brain age, achieving a remarkable mean absolute error of 2.12 years in the UK Biobank. The results establish our approach as state-of-the-art compared to other competing brain age estimation models. In addition, the instance contribution scores identify the varied importance of brain areas for aging prediction, which provides deeper insights into the understanding of brain aging.
\end{abstract}

\end{frontmatter}

\section{Introduction}
\label{sec:introduction}
Aging is commonly defined in most studies as a temporal decline in function that affects a wide range of organisms~\cite{lopez2013hallmarks, adler2007motif, cole2019brain}. While great efforts have been made towards understanding aging, the complex biology of the aging process remains to be fully characterised~\cite{lopez2013hallmarks}. As individuals age, cognitive decline becomes more prevalent, increasing the risk of neurodegenerative disorders and dementia. However, the aging process itself is hardly observable directly. Instead, it can often be inferred from quantifiable accompanying phenomena. In the field of biogerontology, researchers have long been searching for these quantifiable features, known as biomarkers of aging~\cite{cole2019brain}. It's well recognized that throughout the aging process, there are notable changes in both the morphology and function of tissues and organs within the body~\cite{cole2019brain}. Specifically, in the brain, aging induces several physiological changes such as ventricular enlargement, cortical thinning, and the accumulation of white matter hyperintensities~\cite{terry1987neocortical, fjell2014accelerating, fotenos2005normative}. Consequently, there is an opportunity to use neuroimaging technologies for measuring brain age and capturing structural changes in the brain.

Deep learning techniques have achieved remarkable success in medical image analysis, primarily due to their proficiency in learning comprehensive features from clinical imaging datasets. These methods have been applied in a wide range of areas, including object segmentation, disease diagnosis, and brain age estimation~\cite{he2021multi, chen2023brain}. There has been an increasing interest on estimating brain age using deep learning approaches on structural magnetic resonance imaging (sMRI)~\cite{liu2017deep, madan2016cortical, liem2017predicting, lin2016predicting, xu2021brain, cherubini2016importance, feng2020estimating, bashyam2020mri, dinsdale2021learning, he2021multi, lam2020accurate, he2022deep, beheshti2021predicting, ganaie2022brain}. sMRI is capable of capturing changes in brain anatomy and morphological atrophy. Compared to other biomedical imaging techniques, sMRI allows non-invasive examination of a wide range of structural and functional aspects of the human brain, making it a widely utilized tool in both brain aging research and clinical studies of brain disorders~\cite{cherubini2016importance, feng2020estimating, bashyam2020mri, dinsdale2021learning, he2021multi}. Typically, individuals without neurological disorders are assumed to have a normal brain aging trajectory, resulting in an approximate congruence between their brain age and chronological age. To achieve accurate prediction of brain age, it is common practice to train a deep learning model with sMRI data from healthy individuals, using their chronological age as the target.

Recent advances in brain age estimation have seen the successful exploration of deep learning techniques in MRI analysis. For example, BrainNet~\cite{bashyam2020mri} independently predicts an individual's age using 2D slices derived from 3D sMRI, without considering the relationships between slices on age prediction. Recurrent CNN~\cite{lam2020accurate} takes the extracted slice features by a convolutional neural network (CNN) as input to a recurrent neural network. It learns the inter-slice information but fails to provide interpretability of each slice's contribution to the age prediction. In another line of study, Graph Transformer~\cite{cai2022graph} is proposed, which constructs brain networks to learn regional geometric information and obtain age-related features specifically within local regions of interest (ROI) for age prediction. However, it does not take into account the global information in non-ROI regions, thereby possibly compromising the accuracy of the overall prediction. In addition, a Simple Fully Convolutional Network (SFCN)~\cite{peng2021accurate} employs a 3D CNN framework to capture whole image features for age prediction and yields improved estimation results.

While deep learning-based models have greatly enhanced the performance of brain age estimation, many models using the entire 3D MRI as input may ignore the inhomogeneity in aging across different brain areas, making the network focus solely on salient information and thus potentially ignoring finer details~\cite{cole2017predicting}. Recent evidence suggests that age-related metabolism is most pronounced in the frontal lobes, posterior cingulate, and posterior parietal lobes, while the thalamus, putamen, and pallidum are less susceptible to age-related changes~\cite{lee2022deep, kaufmann2019common, zhu2023investigating}. In addition, Man et al. report that brain age is associated with brain volume, including subcortical regions and the prefrontal cortex~\cite{zhu2023investigating}. These prior studies indicate that some regions exhibit significant age-related changes, while others show relatively minor. In other words, different regions contribute to the aging process at different degrees. On the contrary, many models rely on 2D slices as input. However, they treat each slice independently and average the model outputs for prediction which overlooks the inter-slice relationships and often yields suboptimal results. Therefore, it is critical to build a model for capturing the unique patterns of aging across the brain with sufficient consideration of inter- and intra-slice features. Additionally, sMRI can contain a wealth of anatomical and morphological information, while only a portion of them is directly relevant to aging. The redundant information may interfere with the estimation of age, leading to degraded performance.

Based on the above observations, we propose a novel end-to-end architecture called Dual Graph Attention-based Disentanglement Multiple Instance Learning (DGA-DMIL) specifically designed for accurate brain age estimation. It takes multiple 2D slices in sMRI as input and uniformly groups them as a bag of instances, which is fed into a 2D CNN backbone module to capture spatial detail within each instance. A dual graph attention aggregator is then proposed to combine the spatial backbone features of inter- and intra-instances. The aggregator is capable of weighing different contributions of instances to the brain aging process, thereby addressing the issue of brain heterogeneity in aging. Furthermore, we introduce a disentanglement branch to separate age-related features from age-independent features to prevent the negative impact of redundant information on age prediction. The effectiveness of the proposed model is extensively validated on two datasets, UK Biobank and ADNI, with a total of 35,388 samples which is one of the largest sample sizes for brain age estimation. It achieves a mean absolute error (MAE) of 2.12 years in UK Biobank and 2.81 years in ADNI, surpassing the performance of existing state-of-the-art approaches. In addition, our method proves valuable in identifying slices that significantly contribute to brain age estimation, facilitating the localization of the age-related regions. In contrast to prior studies, the main contributions of this work can be summarised as follows:

\begin{figure*}
    \centering
    \includegraphics[width=1\linewidth]{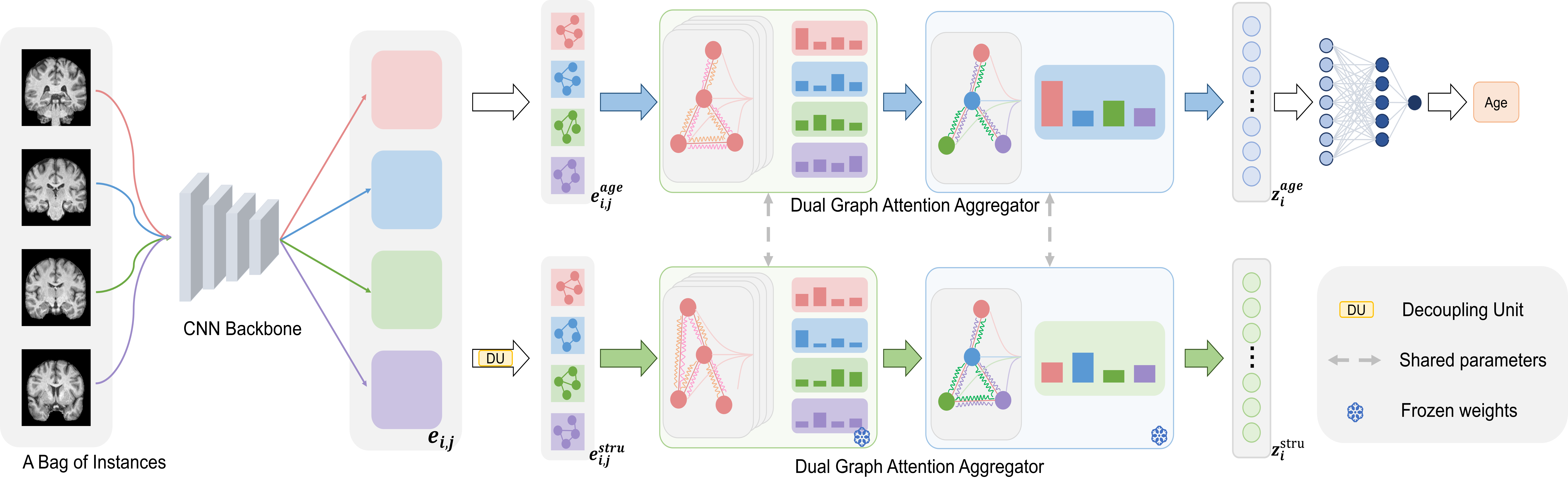}
    \caption{The framework of the proposed DGA-DMIL. A bag of instances is fed into the proposed DGA-DMIL network for predicting age accurately. The essence of DGA-DMIL lies in two key components: (1) Dual Graph Attention Aggregator, responsible for aggregating both intra-instance and inter-instance features (The green box indicates the spatial aggregator $A_{\theta, S}$ and the blue box indicates the instance aggregator $A_{\theta, I}$), and (2) Disentanglement Branch, which separates the backbone features into age-related and age-irrelevant features. (Blue arrows symbolize age-related branches, whereas green arrows signify decoupled branches).}
    \label{fig:DGA_DMIL}
\end{figure*}

\begin{itemize}
     \item We propose a novel end-to-end multi-instance learning framework, called DGA-DMIL, which can fully exploit intra- and inter-instance relationships for accurate brain age estimation.
    \item To address the challenge of heterogeneous aging, we introduce a dual graph attention aggregator to weigh the contribution of different instances of the aging process and to aggregate instance features. In addition, we use contribution scores to visualize the impact of different regions on aging.
    \item To ameliorate the influence of redundant information, we propose a disentanglement branch that separates age-related and age-independent features to prevent the adverse effect of redundant information on age prediction.
    \item Substantial experiments are conducted to verify the effectiveness of our proposed method. Remarkably, our method outperforms all existing state-of-the-art approaches, demonstrating exceptional estimation performance with a Mean Absolute Error (MAE) of 2.12 years in the UK Biobank. 
\end{itemize}

The rest of the paper is organized as follows. In Section~\ref{sec:related}, we introduce the related works, followed by the introduction of the proposed method in detail in Section~\ref{sec:method}. In Section ~\ref{sec:experiments} and~\ref{sec:results}, the experiments and the results are presented, respectively. Finally, we discuss and conclude the paper in Section~\ref{sec:discussion} and~\ref{sec:conclusion}, respectively.

\section{RELATED WORK}
\label{sec:related}

\subsection{Brain age estimation}
With superior feature representation abilities, deep learning-based techniques have been successfully applied to medical image analysis in various medical fields including brain age estimation~\cite{cherubini2016importance, feng2020estimating, bashyam2020mri, dinsdale2021learning, he2021multi, kassani2019multimodal, hu2019hierarchical}. Most existing brain age estimation methods rely on CNN or Transformer architectures to extract features from 2D MRI slices or 3D volumes. One notable approach adopts the two-stage age network (TSAN) to estimate a rough age at the first stage, followed by a second stage that refines the estimation for improved accuracy~\cite{cheng2021brain}. This approach achieves an MAE of 2.43 years on a dataset of 6,586 samples. Another lightweight network, SFCN, proposes an architecture without fully connected layers in brain estimation~\cite{peng2021accurate} and achieves an MAE of 2.14 years on a larger dataset of 14,503 samples. In addition, a global-local transformer model is introduced to capture both global contexts from the entire slices and local fine-grained features from specific patches~\cite{he2021global}. By using attention mechanisms to combine these features, it achieves an MAE of 2.70 years on a dataset of 8,379 samples.

To the best of our knowledge, our approach is the first attempt to use the MIL method for age prediction based on brain sMRI data. Notably, in this study, our proposed method allows us to exploit the attention scores within and between multiple brain slices to estimate brain age. In addition, we introduce a decoupling branch to effectively separate age-related features from age-independent features, further improving the performance of age prediction. Our method achieved an MAE of 2.12 years on the UK Biobank dataset of 35,291 individuals, outperforming existing approaches.

\subsection{Multi-instance learning}
Multi-instance learning (MIL) is a type of weakly supervised learning in which labels are only assigned to a bag of instances, while the labels of instances within the bag remain unknown~\cite{zhou2018brief}. MIL can accurately predict the label of the bag by taking into account the effect of the instance on the final result and provide insights on the contributions of intra- and inter-instance~\cite{han2020accurate, chikontwe2021dual}. Thus, it has been widely applied in various domains, including medical imaging~\cite{shao2021transmil}. Ilse et al. propose an attention-based MIL pooling mechanism focusing on cancer detection in histopathological images~\cite{ilse2018attention}. Chikontwe combined MIL with contrastive learning to identify discriminative features for rapid diagnosis of COVID-19~\cite{chikontwe2021dual}. Marx et al. apply MIL to predict age based on the hippocampal slices~\cite{marx2023histopathologic}. However, it is limited to the hippocampal region, neglecting important information in other brain areas.


\subsection{Decoupled Representation Learning}
Disentangled representation learning aims to capture the explanatory factors from diverse data variations, and has attracted considerable attention in recent years. It has shown great promise in the field of computer vision~\cite{kinney2014equitability, hou2021disentangled}. Previous studies exploit annotated data to separate representations into task-relevant and task-irrelevant features~\cite{hou2021disentangled}. One notable approach, InfoGAN~\cite{chen2016infogan}, maximizes the mutual information between hidden and data variables to successfully disentangle latent factors and thus improve performance. Recently, Hu et al.~\cite{hu2020disentangled} introduce the adversarial autoencoder framework to decouple shared and complementary features across different modalities for accurate brain age prediction. There are notable differences in our proposed method with~\cite{hu2020disentangled}. (1) Architecture: We adopt an end-to-end non-generative network framework. 
(2) Decoupling goal: Our research focuses on decoupling age-related and age-independent features, which is particularly beneficial for brain age estimation.

\section{Method} \label{sec:method}

\subsection{Overview}
\subsubsection{Problem Formulation}
In traditional classification or regression tasks, the goal is to build a model that can accurately predict the target variable, denoted as $y\in R$, given an instance represented by $x\in R^{D}$~\cite{ilse2018attention}. However, in the context of MIL, the input to the model is a collection of instances called a bag, which is represented as $X_i = \left\{ x_{i,1}, x_{i,2}, ..., x_{i,k} \right\} \in R^{K*D}$, where K is the number of instances within the bag, and D represents the dimension of each instance. There exists a label $y \in R$ at the bag level. The main goal is to gain insight into the inherent properties and contributions of individual instances within a collection of instances for the overall label assigned to the collection. Ultimately, the model aims to make predictions about the label $y$ for the given collection of instances. The formulation of MIL for a regression task is expressed as,

\begin{equation}\hat{y_{i} } = S(X_{i} )\in R^{D}.\label{eq1}\end{equation}

Here S denotes the scoring function, i.e. the MIL model, $\hat{y_{i} }$ is the prediction. In our task, we use a brain age estimation dataset denoted as $D = \left \{ \left ( X_i, y_i \right )_{i=1}^{n} \right \} $, consisting of a set of labeled 3D MRI scans $X_i \in R^{H*W*D}$ and age labels $y_i \in R$ sampled from a joint distribution defined as $\mathcal{X\times Y} $. We combine m adjacent slices into one instance given the similarities between adjacent brain MRI slices. Each instance is considered as a 2D image with dimensions H*W and m channels as input to our network. Based on experimental analysis, the optimal value for m is set to be 3. In the context of an sMRI scan, we choose the collection of non-overlapping 2D instances as a bag.

\subsubsection{Workflow}
In this study, we propose a new end-to-end deep learning network called DGA-DMIL for predicting the brain age of an individual using structural MRI data. Figure~\ref{fig:DGA_DMIL} is an illustration of our proposed framework. Initially, a bag of instances is fed into the feature extractor denoted as $E$, which utilizes a CNN backbone pre-trained on ImageNet. It generates a spatial feature map for each instance, represented as $e_{i,j}=E(x_{i,j})\in R^{C*H*W}$, where C, H, and W correspond to the number of channels, height, and width of features, respectively. We then propose a dual graph attention aggregator, consisting of a spatial aggregator $A_{\theta, S}$ and an instance aggregator $A_{\theta, I}$. The spatial aggregator, $A_{\theta, S}$, combines features from deep spatial feature maps within each instance. Subsequently, the instance aggregator, $A_{\theta, I}$, aggregates features from deep representations of multiple instances within a bag. Furthermore, we propose a disentanglement branch to decouple age-related features from age-independent information. The obtained backbone features are then fed into the decoupling unit to generate age features and structural information. At the instance level, a loss function is proposed to encourage structural information from different individuals to be close to each other. At the bag level, an additional loss is introduced to ensure independence between two sets of features. Finally, the age-related representations are employed to estimate the brain age. The details of the framework are as follows. 

\begin{figure}
    \centering
    \includegraphics[width=\linewidth]{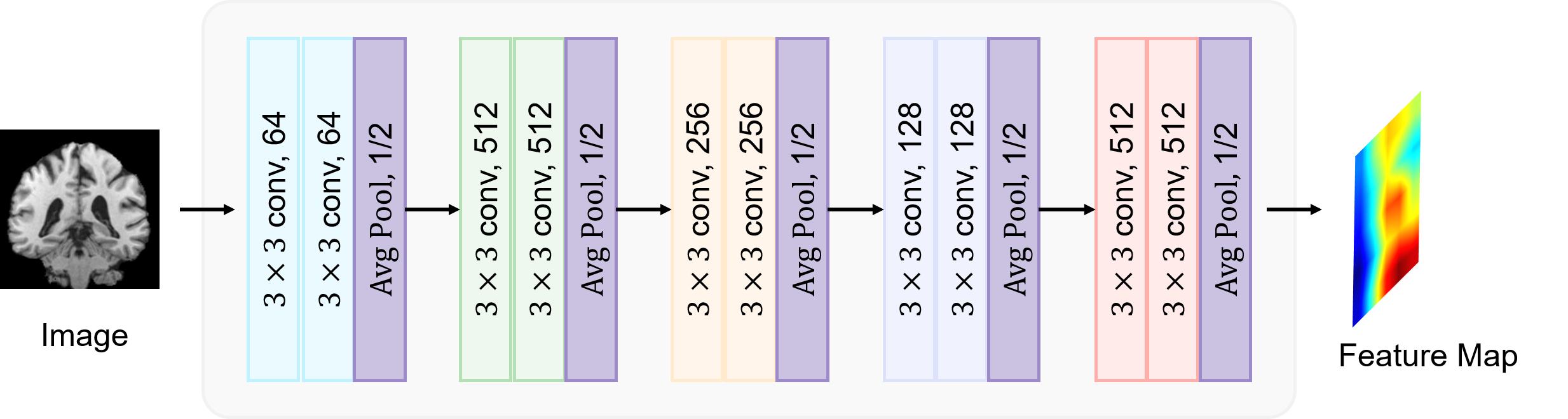}
    \caption{The backbone of the convolutional neural network is responsible for taking a brain image as input and converting it into a deep feature representation. The network comprises 10 blocks, each consisting of a convolutional layer, a batch normalization layer, and a ReLU layer. The spatial resolution is reduced through the max-pooling layer.}
    \label{fig:VGG_Backbone}
\end{figure}

\subsection{CNN backbone}
To analyze sMRI data, we adopt a CNN network as the underlying architecture to extract initial features from instances, as shown in Figure~\ref{fig:VGG_Backbone}. To prevent potential overfitting issues associated with deep networks, we employ the VGG13\cite{simonyan2014very} model as our CNN backbone, which is a well-established architecture widely used in medical imaging. The VGG13 model consists of eight blocks, each comprising a $3 \times 3$ convolutional layer with 1 padding, a batch normalization layer, and a ReLU activation layer. After every two blocks, a maximum pooling layer with a $2 \times 2$ kernel size and a stride of 2 is applied to gradually reduce the size of the spatial feature map. The number of channels utilized in each block is $[64, 128, 256, 512]$.

Considering that the size of natural image datasets is typically larger than that of medical image datasets, pre-trained models on natural images tend to exhibit favorable performance in medical imaging tasks. Therefore, we employ a pre-trained VGG13 model on ImageNet, as our feature extraction model. Following this, we incorporate two convolutional blocks that include a $1 \times 1$ convolutional kernel, a batch normalization layer, and a ReLU activation layer. These additional blocks facilitate the integration of features across channels. The backbone network transforms the input instances into deep features for capturing high-level characteristics.

\subsubsection{Graph Attention Mechanism}
\begin{figure}
    \centering
    \includegraphics[width=\linewidth]{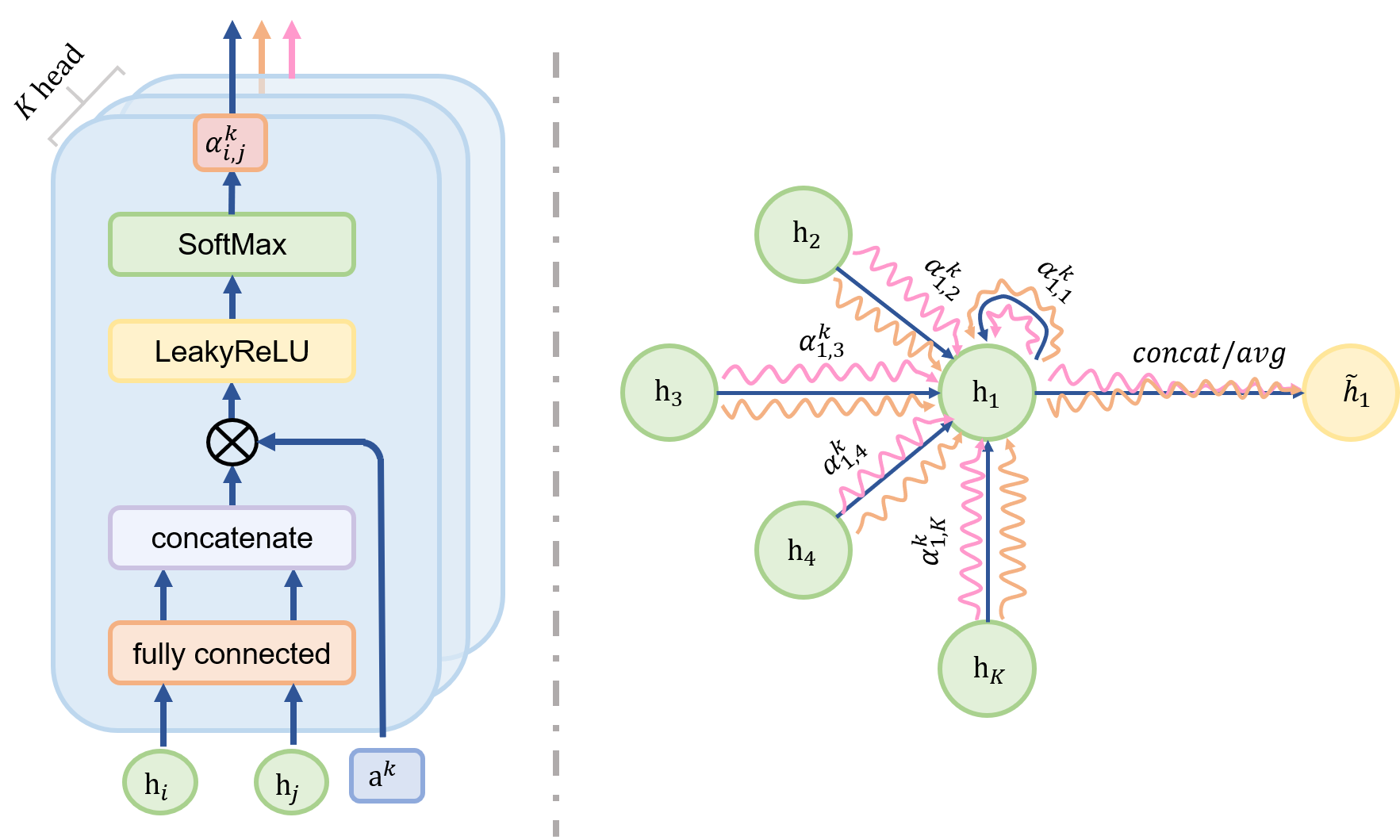}
    \caption{On the left, our model employs an attention mechanism, characterized by a fully connected layer and a weight vector $a^k$, incorporating a LeakyReLU activation and SoftMax function. On the right, an illustration depicts multi-head attention, featuring K = 3 heads, executed by node 1 within its neighborhood. Various arrow styles and colors signify distinct attention computations. The features from each head are combined, concatenated, and averaged to derive $\tilde{h}_1$.}
    \label{fig:Graph_Attention}
\end{figure}

The Graph Attention (GAT) assigns varying weights to neighboring nodes, enabling effective aggregation of neighborhood features. As a result, GAT can capture global information of the entire graph, thereby further enhancing the overall performance of information aggregation, as shown in Figure~\ref{fig:Graph_Attention}. The mechanism for attention scores within the GAT layers can be outlined as follows.

\begin{equation}
\begin{aligned}
v_i^{k} &= W^k h_i \\
e_{i,j}^{k} &= LeakyReLU(a^{k} \left [v_i^{k} ||v_j^{k}\right ])
\end{aligned}
\label{eq2}\end{equation}

\noindent where $h_i$ represents features of neighboring nodes, $W^k \in R^{d_k * d}$ is the weight parameter for improving the expression of node features, and $a^{k} \in R^{2*d_k}$ is the weight parameter of the attention mechanism. The attention score between nodes j and i, denoted as $e_{i,j}$, is computed using the graph attention mechanism, where two neighboring nodes' features are concatenated and multiplied by a linear weight parameter $a^{k}$. It employs a computationally resource-efficient method to calculate the attention score, which differs from the approach used in Transformers. $K$ denotes the number of graph attention heads, and increasing its value enhances the parallelism of computation. However, it also leads to a higher computational effort. In this paper, we select K to be 8. The GAT layer uses the obtained attention scores to aggregate the features of neighboring nodes.

\begin{equation}
\begin{aligned}
\alpha _{i,j}^{k} &= Softmax_{j\in N_i } (e_{i,j}^{k}) \\
\tilde{h}_{i} &= \left | \right | _{k=1}^{K} \sum_{j\in N_i} \alpha _{i,j}^{k} v_j^{k}
\end{aligned}
\label{eq3}\end{equation}

\noindent where $N_i$ represents the neighbors of node i. For node i, the features of the neighboring nodes are aggregated to node i through the weights $\alpha _{i,j}$, and the outputs from different attention heads are concatenated. This process allows for the pooling of features from different nodes in the graph using the multi-head graph attention mechanism. Then, the GAT layer incorporates residual connections and a Feed-Forward Network (FFN). The residual connections ensure a reliable flow of gradients during backpropagation, while Layer Normalization (LN) stabilizes the output distribution and accelerates convergence. The FFN is capable of nonlinear fitting, which improves information aggregation for MIL pooling.

\begin{figure*}
    \centering
    \includegraphics[width=0.8\linewidth]{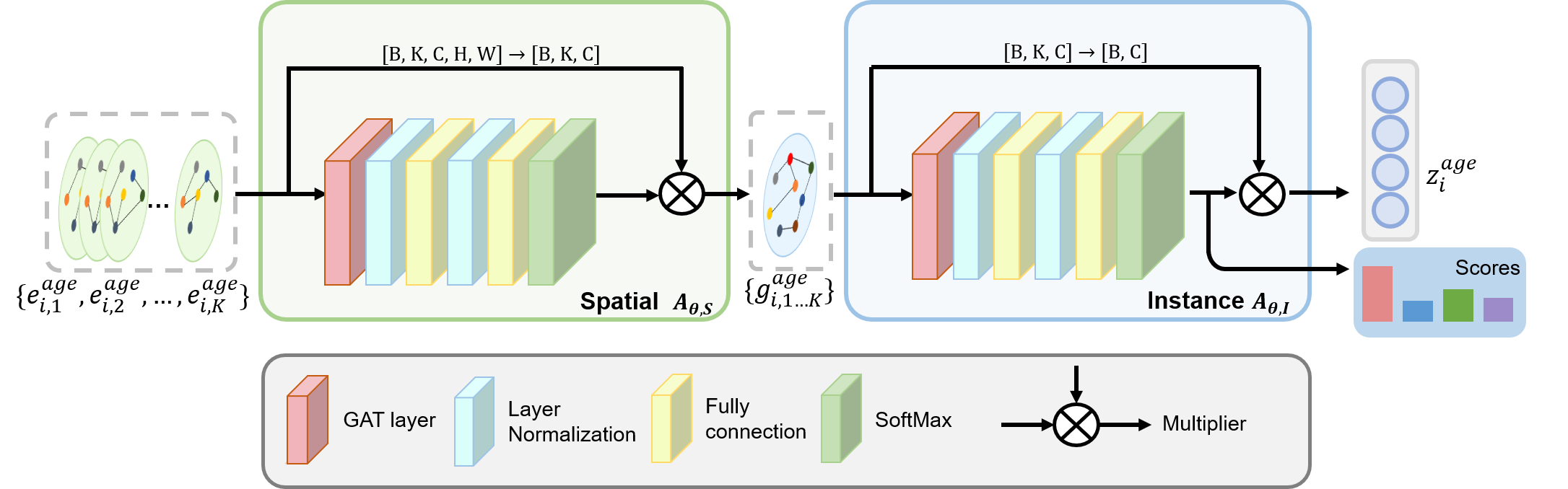}
    \caption{The illustration of the Dual Graph Attention Aggregator Block: The aggregators \(A_{\theta, S}\) specialize in aggregating spatial information from the deep feature map within an instance. Subsequently, \(A_{\theta, I}\) is designed to aggregate instance information from the feature set within a bag.  The Dual Graph Attention Aggregator aims to obtain a representative for brain age estimation. In addition, instance scores are output to evaluate the impact of individual instances on brain age estimation.}
    \label{fig:Dual_Aggregator}
\end{figure*}

\subsubsection{Dual Graph Attention Aggregator}
Inspired by the attention-based MIL pooling method developed in~\cite{ilse2018attention}, we propose a novel MIL module, the dual graph attention aggregator (as illustrated in Figure~\ref{fig:Dual_Aggregator}), building on the graph attention mechanism. The dual graph attention aggregator consists of a spacial-level feature aggregation network $A_{\theta, S} (R^{K*C*H*W} \to R^{K*C})$ and an instance-level feature aggregation network $A_{\theta, I} (R^{K*C} \to R^C)$. In the aggregator $A_{\theta, S}$, our goal is to aggregate spatial features from the feature map $e_{i,j}$ into channel features. By passing the feature map $e_{i,j}$ through the GAT block, which consists of GAT, LN, and full connection (FC) layers to reduce the channel dimension to 1, and a softmax layer for normalization, we obtain the final spatial score $g_{i,j}$. To compute the spatial score $g_{i,j} \in R^{1*H*W}$, we perform an element-wise multiplicative convolution of each branch, taking the output layers of the previous layer as input. Mathematically, $g_{i,j}^{'} = g_{i,j}*e_{i,j} \in R^{C}$, which we denote as $g_{i,j}$ for consistency. In particular, our approach differs from simple global average pooling of the initial features as we adopt the graph attention mechanism to assign different weights within the spatial feature map.

Once we have obtained the deep semantic features $g_{i,j}$ from a bag of instances, we use a simple but effective method to generate a graph to obtain the bag-level features. This involves calculating the cosine similarity between each node and selecting the n nodes with the lowest similarity values as edges. These edges determine the flow of feature information between the nodes. In our experiments, we set the value of n to 8. To perform the aggregation of instance-level features to bag-level features, we use the aggregator $A_{\theta, I}$. The goal of the aggregator $A_{\theta, I}$ is to aggregate the deep semantic features of instances. We adopt the same structural design as previously applied to the initial backbone feature map $e_{i,j}$. Formally, let $\left \{g_{1},g_{2},...,g_{K} \right \}$ represent the features of instances with a bag.  The graph attention based MIL is defined as:

\begin{equation}
z = \sum_{i=1}^{K} s_{i}*g_{i}
\label{eq4}\end{equation}

\noindent with,

\begin{equation}
s_{i} = \frac{exp(w^{T}*GAT(g_{i})) }{ {\sum_{j=1}^{K}exp(w^{T}*GAT(g_{j}))} } 
\label{eq5}\end{equation}

\noindent where $w \in R^{K*1}$ is the trainable parameter. The main difference with existing attention mechanisms is that we use the graph attention mechanism, which allows information exchange between similar instances and facilitates effective information aggregation. In addition, the attention weights assigned to each instance indicate its relevance to the prediction of the whole bag. A deep 3D instance is considered critical if it receives a higher attention weight. Consequently, the attention weights provide insight into the contribution of each instance to the bag, improving the interpretability of the MIL model. The bag-level age features are then fed into a classifier network $H$ to obtain the predicted age of an individual, denoted as $\hat{y_i} = H(z)$. Formally, we use the Mean Squared Error (MSE) loss to encourage more accurate age prediction.
\begin{equation}{L_{MSE} } = \frac{1}{N}  {\textstyle \sum_{i}(\hat{y_i} - y_i)^2} \label{eq6}\end{equation}

\subsection{Disentanglement Branch}
In the context of characterizing structural MRI, two components are entangled: age-related features and age-independent structural features. Therefore, we propose a disentanglement branch, aiming at decomposing the features extracted from the backbone into two sets. This separation helps to obtain age-related information, thus reducing the influence of redundant information on the age estimation. To achieve this, we use a linear decomposition technique for simplicity. A decoupling unit, consisting of two CNN layers with $1 \times 1$ kernels and an FC layer, is applied to the initial feature map and extracts structural features. The age features are then derived by subtracting the initial feature map from the structural features.

\begin{equation}
e_{i,j}^{age}=e_{i,j}-e_{i,j}^{stru},\,\,\,\,\,\,\,e_{i,j}^{stru}=\Psi (e_{i,j})
\label{eq7}\end{equation}

To ensure that the spatial feature $e_{i,j}^{age}$ extracted from each instance $x_{i,j}$ primarily captures age-related information rather than structural details, we use an MSE loss function to guide the age feature extraction network. The constraint helps to focus on age-related information during feature extraction.

\begin{equation}
L_{MSE0}=\frac{1}{N} {\textstyle \sum_{i} (y_i - y_i^0 ) ^2},\,\,\,\,\,\,\,\hat{y_i^0}= \varphi (e_{i,j}^{age}) 
\label{eq8}\end{equation}

The initial age is obtained by global average pooling and MLP prediction, to minimize the loss between the predicted age and the true age, thereby encouraging the extracted semantic features to be more age-related. Since structural features tend to show similarity across individuals, we include constraints to maximize the cosine similarity of structural features $e_{i,j}^{stru}$ at the instance level to ensure that the extracted features characterize the structural information.

\begin{equation}
\begin{aligned}
L_{decp1} &= -\frac{1}{N} {\textstyle \sum_{i=1}^{N}l_i} \\
l_i &= \frac{1}{K} {\textstyle \sum_{j=1}^{K}1_{i\ne k}cos(e_{i,j}^{stru},e_{k,j}^{stru})}
\end{aligned}
\label{eq9}\end{equation}

The value of k is randomly chosen from the range of $[1, N]$, corresponding to the batch size. To increase the effectiveness of the decoupling, we also apply decoupling at the bag level, which further constrains the extracted features. Specifically, we feed the decoupled $e_{i,j}^{stru}$ and $e_{i,j}^{age}$ into two aggregator networks with common parameters, resulting in bag-level features $z_i^{stru}$ and $z_i^{age}$, respectively. Based on prior assumptions, at the bag level, we expect these two sets of features to have minimal correlation. To enforce this, we minimize the cosine similarity between them.

\begin{equation}
L_{decp2} = \frac{1}{N} {\textstyle \sum_{i=1}^{N}cos(z_{i}^{stru},z_{k}^{age})}
\label{eq10}\end{equation}

\subsection{Overall Loss Function}
The proposed brain age estimation framework can be trained in a supervised manner. The overall objective $L$ is as follows:

\begin{equation}
L = L_{MSE} + \lambda _2 L_{MSE0} + \lambda _3 L_{decp1} + \lambda _4 L_{decp2}
\label{eq11}\end{equation}

\noindent where $L_{MSE0}$ and $L_{MSE}$ denote losses for preliminary and final brain age prediction. $L_{decp1}$ and $L_{decp1}$ denote the decoupling loss for mitigating the effects of redundant information. $\lambda _2$, $\lambda _3$, and $\lambda _4$ are trade-off parameters controlling the importance of each component.

\begin{table}[]
\caption{Demographic information of the studied subjects from ADNI and UKB datasets.}
\centering
\resizebox{\linewidth}{!}{%
\begin{tabular}{lllll}
\toprule
Dataset   & Number & Age Range & Age Statistics & Male/Female \\ \midrule
UK Biobank & 35291  & 44-82     & 63.8 $\pm$ 7.68      & 16789/18502 \\
ADNI      & 127     & 60-90   & 76.3 $\pm$ 5.14      & 62/65       \\ \bottomrule
\end{tabular}}
\label{tab1}
\end{table}

\section{EXPERIMENTS} \label{sec:experiments}

\subsection{Datasets and Preprocessing}
The proposed method is validated in this study on two publicly available datasets, UK Biobank (UKB)~\footnote{\url{https://biobank.ctsu.ox.ac.uk/}} and ADNI~\footnote{\url{http://adni.loni.usc.edu/}}. The demographic information of the subjects from both datasets is provided in TABLE~\ref{tab1}. For the UKB dataset, we select a subset of 35,291 individuals with neuroimaging data who have no evidence of neurological pathology and no psychiatric diagnosis according to ICD-10 criteria. In addition, our study includes the collection of structural MRI (sMRI) data from a total of 127 healthy subjects from the ADNI dataset.

A standard preprocessing pipeline using FMRIB Software Library (FSL, \href{https://fsl.fmrib. ox.ac.uk/}{https://fsl.fmrib. ox.ac.uk/})~\cite{woolrich2009bayesian, smith2004advances, jenkinson2012fsl} is applied to all sMRI data. It includes several steps such as nonlinear registration~\cite{andersson2007non} to the MNI152 standard space, brain extraction~\cite{smith2002fast, jenkinson2005bet2}, and normalization of voxel values. To ensure consistency of input sizes across different datasets, all pre-processed sMRI images are cropped based on a generated mask that removes uninformative zero-value pixels from the background. As a result, all MRIs have a voxel size of $160 \times 192 \times 160$ and an isotropic spatial resolution of 1 $mm^3$.

\subsection{Experimental Settings}
The proposed method is implemented on four NVIDIA GeForce RTX 3090 24 GB GPUs and the PyTorch package in Python. To assess performance and generalization, experiments are performed on pre-processed T1-weighted sMRI data from the UKB and ADNI datasets, as described in TABLE~\ref{tab1}. For the UKB dataset, the sMRI is randomly divided into 3 non-overlapping sets: a training set (28,675 subjects), a validation set (1,323 subjects), and a test set (5,293 subjects). The training-validation-test split is performed only once and is used in all following experiments. The models that performed best on the validation set are selected for testing. Five-fold cross-validation is then performed on the ADNI dataset. In our experiments, three metrics are used to quantitatively evaluate brain age estimation including mean absolute error (MAE), root mean squared error (RMSE), and Pearson correlation coefficient (PCC). The PCC is calculated between the estimated brain age and chronological age to measure the degree of similarity between the two variables. The average results and their standard deviations are reported. A good estimation is achieved when MAE and MSE are close to 0 and PCC is close to 1.

\subsection{Network Training}
To train the proposed DGA-DMIL network, we use ADAM optimizer\cite{kingma2014adam} for optimization, and initialize it with a learning rate of 0.0001. The batch size is set to be 32 and the disentanglement weights agree $\lambda _2 = 0.1$, $\lambda _3 = 0.05$, and $\lambda _4 = 1$, based on their performance on the validation set. The weights in the CNN layers in the decoupling unit are initialized using the Kaiming weight initialization scheme~\cite{he2015delving}, and the weights in the fully connected layers are initialized with a normal distribution. The bias of the final fully connected layer is set equal to the average age of the training dataset for faster convergence. During training, the sum of the mean squared error loss and the decoupling loss is adopted as the total loss, and backpropagation is performed accordingly. All brain age estimation models are implemented in PyTorch.

For the learning rate schedule, a decay factor of 0.8 is applied if the training loss does not decrease for 5 consecutive epochs. Training is limited to 120 epochs and early termination is applied if the loss does not decrease for 20 consecutive epochs.

\begin{figure}
    \centering
    \includegraphics[width=1\linewidth]{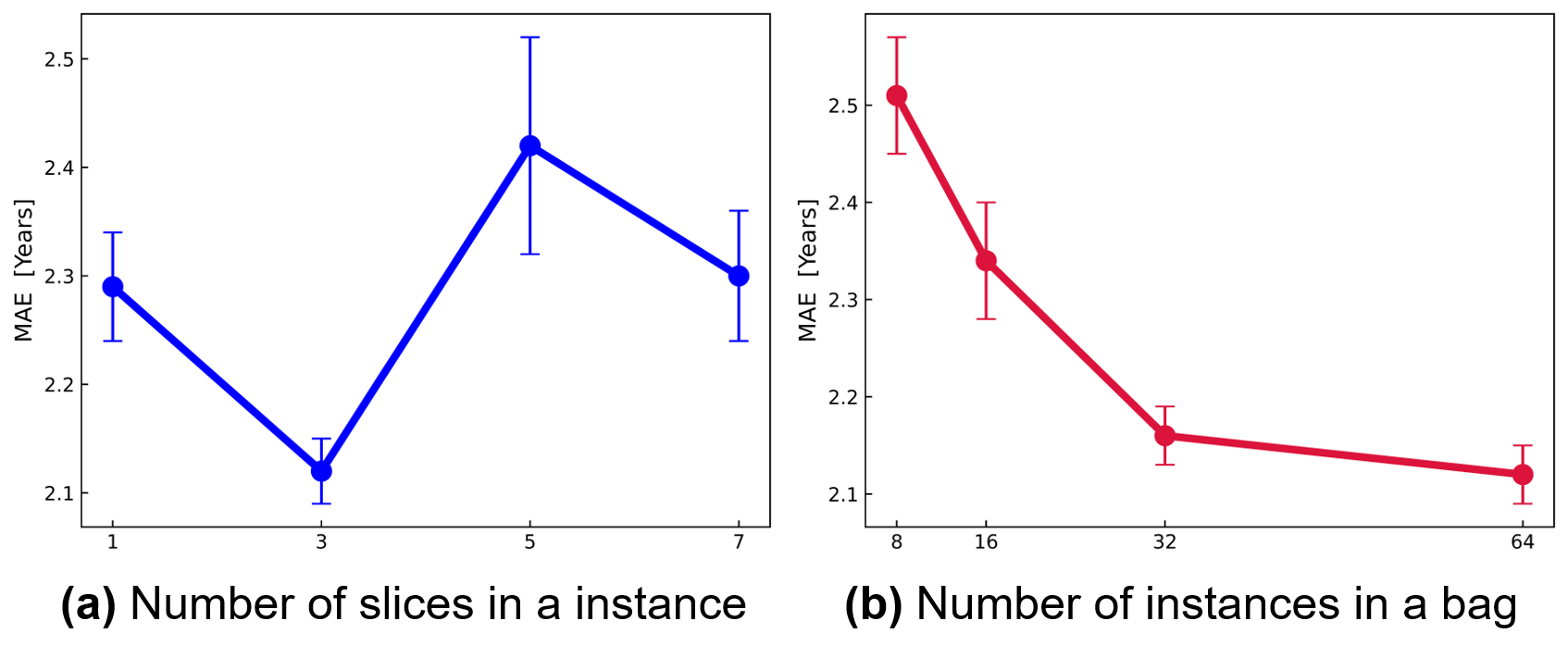}
    \caption{Performance of brain age estimation with different parameters: (a) numbers of slices in an instance; (b) numbers of instances in a bag.}
    \label{fig:enter-label3}
\end{figure}

\begin{figure*}
    \centering
    \includegraphics[width=1\linewidth]{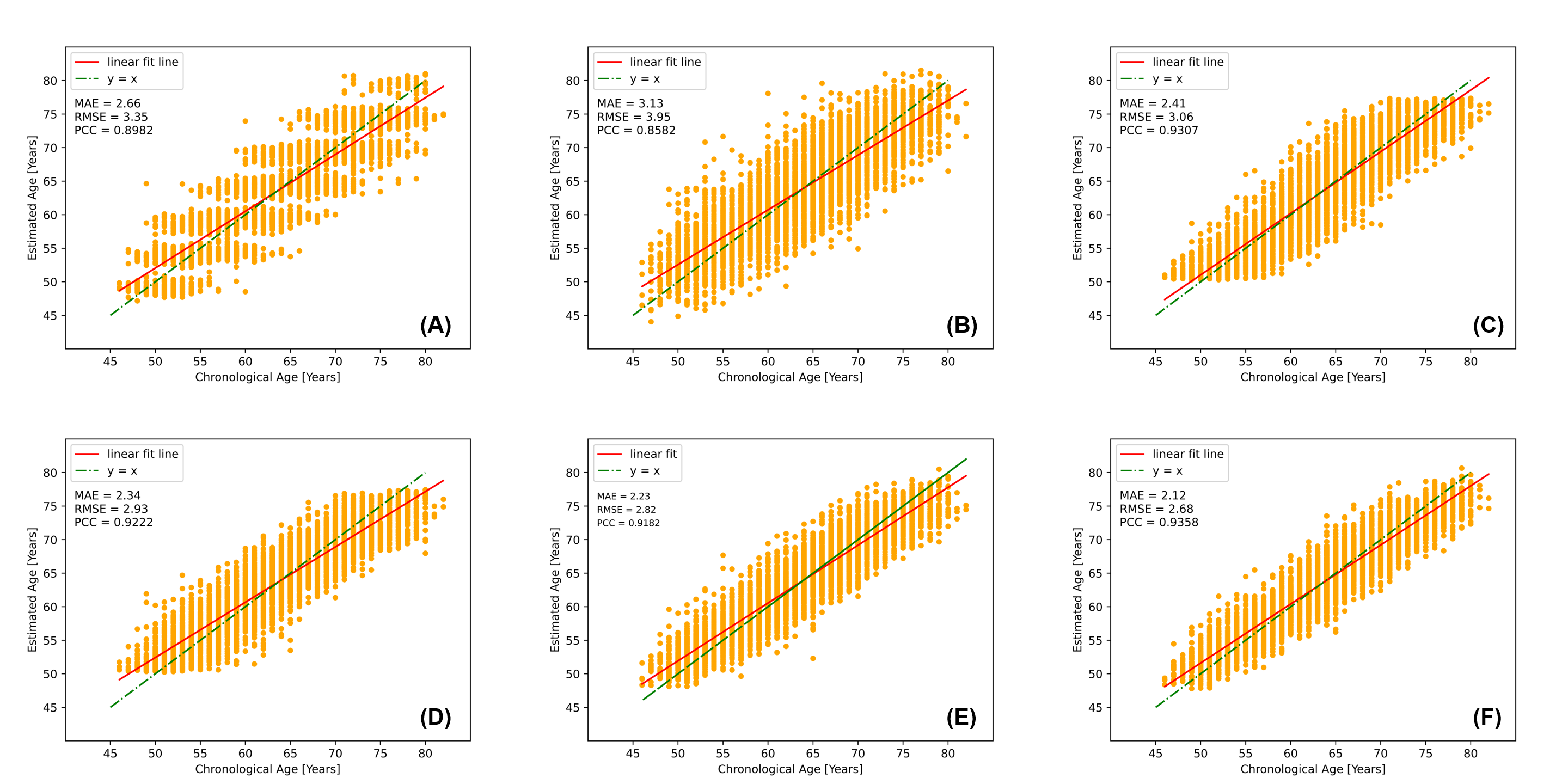}
    \caption{Scatter diagrams of the estimated brain ages and chronological ages by different estimation models. (A) TSAN~\cite{cheng2021brain}; (B) Graph Transformer~\cite{cai2022graph}; (C) Recurrent CNN~\cite{lam2020accurate}; (D) SFCN~\cite{peng2021accurate}; (E) DGA-DMIL (GoogleNet backbone); (F) Our proposed DGA-DMIL framework (VGG backbone). In all figures, the green dashed line indicates the ideal estimation y = x (i.e., the estimated brain age equals the chronological age), and the red line is the linear regression model fitted by the estimated brain age and chronological age. MAE: mean absolute error; RMSE: root means squared error; PCC: Pearson correlation coefficient between the estimated brain age and chronological age.}
    \label{fig:enter-label4}
\end{figure*}

\section{RESULTS} \label{sec:results}
In this section, we investigate the effectiveness of the proposed age estimation framework. It is compared with different baseline models and state-of-the-art architectures. In addition, we perform ablation experiments to validate the proposed graph attention MIL pooling and disentanglement branch. Finally, we visualize the most informative instances for brain age estimation.

\subsection{Parameter Evaluation}
In this framework, the number of slices in an instance (instance size) and the number of instances in a bag (bag size) are two important parameters for estimating brain age. We evaluate the performance of our method by varying the instance size from 1 to 7, as shown in Figure~\ref{fig:enter-label3}(a). It is noted from the figure that the MAE is minimized when the number of slices is set to 3. We conducted experiments with different values from 8 to 64 to investigate the effect of bag size. The maximum bag size we can select is 64, which is affected by the size of the sMRI data. The results are shown in TABLE~\ref{tab2} and Figure~\ref{fig:enter-label3}(b), which indicate that increasing the bag size improves the age prediction performance of the model. Notably, the best results are obtained when k is set to 64. Therefore, in the following sections we fix the instance size to 3 and the bag size to 64.

\begin{table}[]
\centering
\caption{Comparison of varying bag sizes with the proposed method on UKB dataset.}
\resizebox{\linewidth}{!}{%
\begin{tabular}{@{}lccc@{}}
\toprule
\multicolumn{1}{c}{Method} & MAE           & RMSE          & PCC               \\ \midrule
DGA-DMIL (k=8)             & 2.51$\pm$0.06 & 3.16$\pm$0.09 & 0.9093$\pm$0.0014 \\
DGA-DMIL (k=16)            & 2.34$\pm$0.06 & 2.95$\pm$0.08 & 0.9217$\pm$0.0012 \\
DGA-DMIL (k=32)            & 2.16$\pm$0.03 & 2.73$\pm$0.05 & 0.9333$\pm$0.0010 \\
DGA-DMIL (k=64)            & \textbf{2.12$\pm$0.03} & \textbf{2.68$\pm$0.05} & \textbf{0.9358$\pm$0.0011} \\ \bottomrule
\end{tabular}}
\label{tab2}
\end{table}

\begin{table}[]
\caption{Comparison of the brain age estimation results in different backbone and aggregator on UKB dataset.}
\resizebox{\linewidth}{!}{%
\begin{tabular}{@{}lllll@{}}
\toprule
Backbone                       & Aggregator & MAE       & RMSE      & PCC           \\ \midrule
\multirow{2}{*}{ResNet}    & CNN         & 2.18$\pm$0.06 & 2.72$\pm$0.09 & 0.9240$\pm$0.0012 \\
                           & GAT         & 2.14$\pm$0.03 & 2.70$\pm$0.08 & 0.9238$\pm$0.0009 \\ \midrule
\multirow{2}{*}{GoogleNet} & CNN         & 2.27$\pm$0.08 & 2.87$\pm$0.10 & 0.9161$\pm$0.0015 \\
                           & GAT         & 2.23$\pm$0.09 & 2.82$\pm$0.12 & 0.9182$\pm$0.0023 \\ \midrule
\multirow{2}{*}{VGG}       & CNN         & 2.16$\pm$0.06 & 2.68$\pm$0.07 & 0.9347$\pm$0.0010 \\
                           & GAT         & \textbf{2.12$\pm$0.03} & \textbf{2.64$\pm$0.05} & \textbf{0.9358$\pm$0.0011} \\ \bottomrule
\end{tabular}}
\label{tab3}
\end{table}

Once we have obtained the bag of instances, we feed them into a backbone network to extract deep feature maps. We further evaluate the performance of our proposed method based on different CNN backbones, including GoogleNet, ResNet18, and VGG13. As shown in TABLE~\ref{tab3}, the VGG13 backbone achieves the best performance in brain age estimation, which is consistent with the results of a similar study reported in~\cite{he2021global}. To verify the effectiveness of the proposed graph attention-based aggregator, we compare it with a CNN-based aggregator. The CNN aggregator uses two convolution layers and activity layers to reduce the channel of the feature maps. FC layers are then adopted to reduce the channel to 1 for score generation and feature aggregation. Additionally, we compare the effects of different backbones in combination with different MIL pooling strategies. As demonstrated in TABLE~\ref{tab3}, the VGG backbone with graph attention aggregator outperforms all other competing combinations. Therefore, in our experiments, we use VGG as the CNN backbone and GAT as the aggregator.

\subsection{Comparison with Other MIL Models of Brain Age Estimation}
In this experiment, we test the performance of the proposed network on brain age estimation using the UKB dataset. We compare the proposed dual graph attention-based MIL method with other MIL methods, including Deep MIL~\cite{ilse2018attention}, DA-CMIL~\cite{chikontwe2021dual}, and TransMIL~\cite{shao2021transmil}. In our experiments, we employ global mean pooling for the mean operator. Deep MIL adopts a weighted average of instances (low-dimensional embeddings) where the weights are determined by a neural network. DA-CMIL proposes attention-based MIL pooling and contrastive learning to better model the instance. TransMIL introduces a revised self-attention mechanism to reduce the computational complexity of MIL pooling. In addition, TransMIL employs convolution kernels of different sizes in the same layer to encode positional information. The above methods are implemented with our best efforts based on the published codes. These results are provided in TABLE~\ref{tab4}. Our proposed method showed an improvement of 0.33 in the accuracy of brain age prediction. The performance of our proposed MIL model is consistently better than the compared methods.

\begin{table}[]
\centering
\caption{Comparison of the brain age estimation methods using different MIL methods on UKB dataset.}
\resizebox{\linewidth}{!}{%
\begin{tabular}{@{}llll@{}}
\toprule
Method    & MAE           & RMSE          & PCC               \\ \midrule
Deep MIL\cite{ilse2018attention}  & 2.64$\pm$0.18 & 3.30$\pm$0.21 & 0.9092$\pm$0.0066 \\
DA-CMIL\cite{chikontwe2021dual}   & 2.55$\pm$0.10 & 3.18$\pm$0.12 & 0.9278$\pm$0.0056 \\
Trans-MIL\cite{shao2021transmil} & 2.45$\pm$0.16 & 3.08$\pm$0.19 & 0.9138$\pm$0.0044 \\
Proposed Method      & \textbf{2.12$\pm$0.03} & \textbf{2.68$\pm$0.05} & \textbf{0.9358$\pm$0.0011} \\ \bottomrule
\end{tabular}}
\label{tab4}
\end{table}

\subsection{Comparison with SoTA Methods of Brain Age Estimation}
We perform a comprehensive comparison of the proposed method with eight state-of-the-art approaches recently published in the literature, and evaluate them focusing on MAE, MSE, and PCC metrics. The methods compared mainly use deep neural networks, including CNNs, graph-based models, and transformers, for brain age estimation based on T1-weighted sMRI data. To ensure a fair comparison, all methods are implemented with the same training configuration and evaluated on identical training, validation, and test datasets. In this experiment, we implement the public source code of the methods ~\cite{cai2022graph, he2021global, lam2020accurate, cheng2021brain, peng2021accurate} available on GitHub. However, for methods without a PyTorch implementation, such as\cite{bashyam2020mri,lee2022deep,yin2023anatomically}, we made our best efforts to reproduce the methods based on the details provided in their respective papers. It's worth noting that these 3D CNN models follow a standard architecture and are relatively easy to implement.

\begin{table}[]
\centering
\caption{Comparison of our proposed method with SOTA methods on UKB dataset.}
\resizebox{\linewidth}{!}{%
\begin{tabular}{cccc}
\toprule
Method                                                                                         & MAE                    & RMSE                   & PCC                        \\ \midrule
Bashyam et al., 2020~\cite{bashyam2020mri}                & 3.26$\pm$0.20 & 4.05$\pm$0.22 & 0.8604$\pm$0.0029 \\
Lam et al., 2020~\cite{lam2020accurate}            & 2.41$\pm$0.11 & 3.06$\pm$0.14 & 0.9307$\pm$0.0012 \\
He et al., 2021~\cite{he2021global}          & 3.10$\pm$0.23 & 3.85$\pm$0.28 & 0.8515$\pm$0.0161 \\
Cheng et al., 2021~\cite{cheng2021brain}                    & 2.66$\pm$0.05 & 3.35$\pm$0.06 & 0.8982$\pm$0.0022 \\
Peng et al., 2021~\cite{peng2021accurate}                    & 2.34$\pm$0.10 & 2.93$\pm$0.12 & 0.9222$\pm$0.0016 \\
Cai et al., 2022~\cite{cai2022graph}       & 3.13$\pm$0.27 & 3.95$\pm$0.29 & 0.8582$\pm$0.0091 \\
Lee et al., 2022~\cite{lee2022deep}            & 2.61$\pm$0.12 & 3.25$\pm$0.14 & 0.9282$\pm$0.0019 \\
Yin et al., 2023~\cite{yin2023anatomically}          & 2.76$\pm$0.24 & 3.44$\pm$0.29 & 0.9076$\pm$0.0054
\\ \midrule
Proposed Method         & \textbf{2.12$\pm$0.03} & \textbf{2.68$\pm$0.05} & \textbf{0.9358$\pm$0.0011} \\ \bottomrule
\end{tabular}}
\label{tab5}
\end{table}

\begin{table}[]
\centering
\caption{Comparison of our proposed method with SOTA methods on ADNI dataset.}
\resizebox{\linewidth}{!}{%
\begin{threeparttable}[]

\begin{tabular}{cccc}
\toprule
Method                               & MAE                    & RMSE                   & PCC                        \\ \midrule
He et al., 2021~\cite{he2021global}$\diamond$                    & 3.44$\pm$0.53          & 4.56$\pm$0.82          & 0.4169$\pm$0.2162          \\
He et al., 2021~\cite{he2021global}$\ast$                     & 3.32$\pm$0.54          & 4.34$\pm$0.66          & 0.5527$\pm$0.0872          \\ \midrule
Peng et al., 2021~\cite{peng2021accurate}$\diamond$                  & 3.51$\pm$0.55          & 4.52$\pm$0.67          & 0.4545$\pm$0.1489          \\
Peng et al., 2021~\cite{peng2021accurate}$\ast$                    & 3.52$\pm$0.57          & 4.48$\pm$0.57          & 0.4817$\pm$0.1228          \\ \midrule
Lee et al., 2022~\cite{lee2022deep}$\diamond$                   & 3.48$\pm$0.48          & 4.57$\pm$0.37          & 0.4799$\pm$0.1083          \\
Lee et al., 2022~\cite{lee2022deep}$\ast$                     & 3.02$\pm$0.28          & 3.86$\pm$0.28          & 0.6536$\pm$0.0708          \\ \midrule
Proposed Method$\circ$  & 3.58$\pm$0.30          & 4.43$\pm$0.37          & 0.5775$\pm$0.1412          \\
Proposed Method$\diamond$ & 3.22$\pm$0.33          & 4.30$\pm$0.54          & 0.5261$\pm$0.1549          \\
Proposed Method$\ast$   & \textbf{2.79$\pm$0.18} & \textbf{3.61$\pm$0.13} & \textbf{0.6995$\pm$0.1309} \\ \bottomrule
\end{tabular}
\begin{tablenotes}
     \item $\diamond$ denotes the model trained from scratch
     \item $\ast$ denotes the pre-trained model with fine-tuning
     \item $\circ$ denotes the pre-trained model without fine-tuning
\end{tablenotes}
\end{threeparttable}}
\label{tab_adni}
\end{table}

The results are shown in TABLE~\ref{tab5}. On the UK Biobank dataset, it is observed that the 2D CNN method achieves the highest MAE ($>$ 3.20 years) and RMSE ($>$ 4.00 years) among the algorithms compared. This discrepancy could be attributed to the different age-related deterioration in the different 2D slices. Comparatively, both the Recurrent CNN and the Global-Local Transformer methods yield improved results in brain age estimation compared to the 2D CNN models, and the SFCN method outperforms the other 2D CNN methods. However, by integrating representations within and across instances, our proposed method outperforms all other competing models, achieving a further reduction in MAE of 0.22. The superior performance of our proposed method is clearly illustrated in Figure~\ref{fig:enter-label4}, where we compare the brain ages estimated by different brain age estimation models with the chronological age on UK Biobank test data. The standard deviation (STD) of the MAE in Figure~\ref{fig:enter-label4} serves as an indicator of the uncertainty associated with the estimated MAE results. 

\begin{figure}
    \centering
    \includegraphics[width=\linewidth]{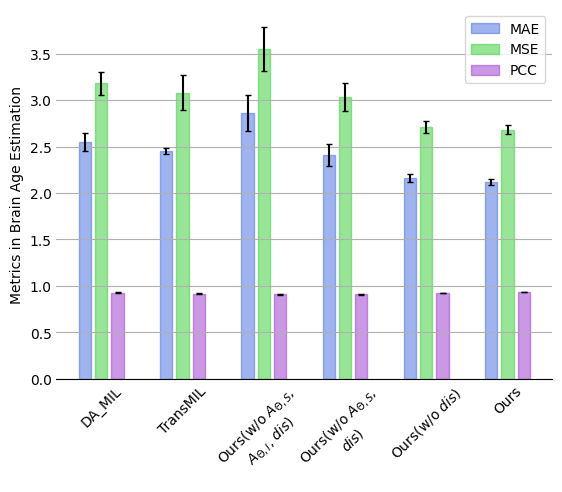}
    \caption{The comparison between multi-instance learning methods and ablation study.}
    \label{fig:enter-label5}
\end{figure}

The results obtained on the ADNI dataset are significantly inferior to those obtained on the UK Biobank dataset, with larger values in MAE and RMSE as well as smaller values in PCC. This discrepancy can be attributed to the significant reduction in data size. Nevertheless, our results still show optimal performance, with a 0.23 improvement over the second-best-performing methods, demonstrating the robustness of our model. It is important to note that the results presented in TABLE~\ref{tab_adni} may differ from those reported in the existing literature due to variations in the data studied and the training configuration.

\subsection{Ablation Experiments of the Framework}
To evaluate the effect of the proposed dual graph attention aggregator and disentanglement branch in the proposed framework, we consider several settings where the dual graph attention aggregator and disentanglement branch are either employed or not (Figure~\ref{fig:enter-label5} and TABLE~\ref{tab_ablation}). Specifically, if the dual graph attention aggregator is excluded, the framework would need to be modified in two aspects; (1) without instance-level MIL pooling of features, we default to using global average pooling (GAP) of instance features for simplicity (denoted as w/o aggregator $A_{\theta, S}$), and (2) without spatial feature aggregation in instances (spatial MIL pooling), one can opt to use the mean of instance features to obtain the overall bag-level feature z (denoted as w/o aggregator $A_{\theta, S}$, $A_{\theta, I}$). It is worth noting that the results without dual aggregator are consistent with the results with mean MIL pooling since we utilize GAP instead of Graph Attention MIL pooling. 

The results demonstrate that the best performance is achieved when the dual graph attention aggregator and the disentanglement branch are adopted. On the other hand, when the attention modules are excluded, a significant drop in overall performance is observed, i.e. 0.74 compared to the best-performing method (TABLE~\ref{tab_ablation}).

\begin{table}[]
\centering
\caption{Ablation studies about different modules of the proposed method on UKB dataset.}
\resizebox{\linewidth}{!}{%
\begin{tabular}{cccc}
\toprule
Method                                                                                         & MAE                    & RMSE                   & PCC                        \\ \midrule
Ours (w/o $A_{\theta,S}$, $A_{\theta,I}$, $dis$) & 2.86$\pm$0.19          & 3.55$\pm$0.24          & 0.9074$\pm$0.0068          \\
Ours (w/o $A_{\theta,S}$, $dis$)                & 2.41$\pm$0.12          & 3.03$\pm$0.15          & 0.9097$\pm$0.0041          \\
Ours (w/o $dis$)                                                                            & 2.16$\pm$0.04          & 2.71$\pm$0.06          & 0.9227$\pm$0.0013          \\
Ours                                                                                       & \textbf{2.12$\pm$0.03} & \textbf{2.68$\pm$0.05} & \textbf{0.9358$\pm$0.0011} \\ \bottomrule
\end{tabular}}
\label{tab_ablation}
\end{table}

\begin{figure*}
    \centering
    \includegraphics[width=1\linewidth]{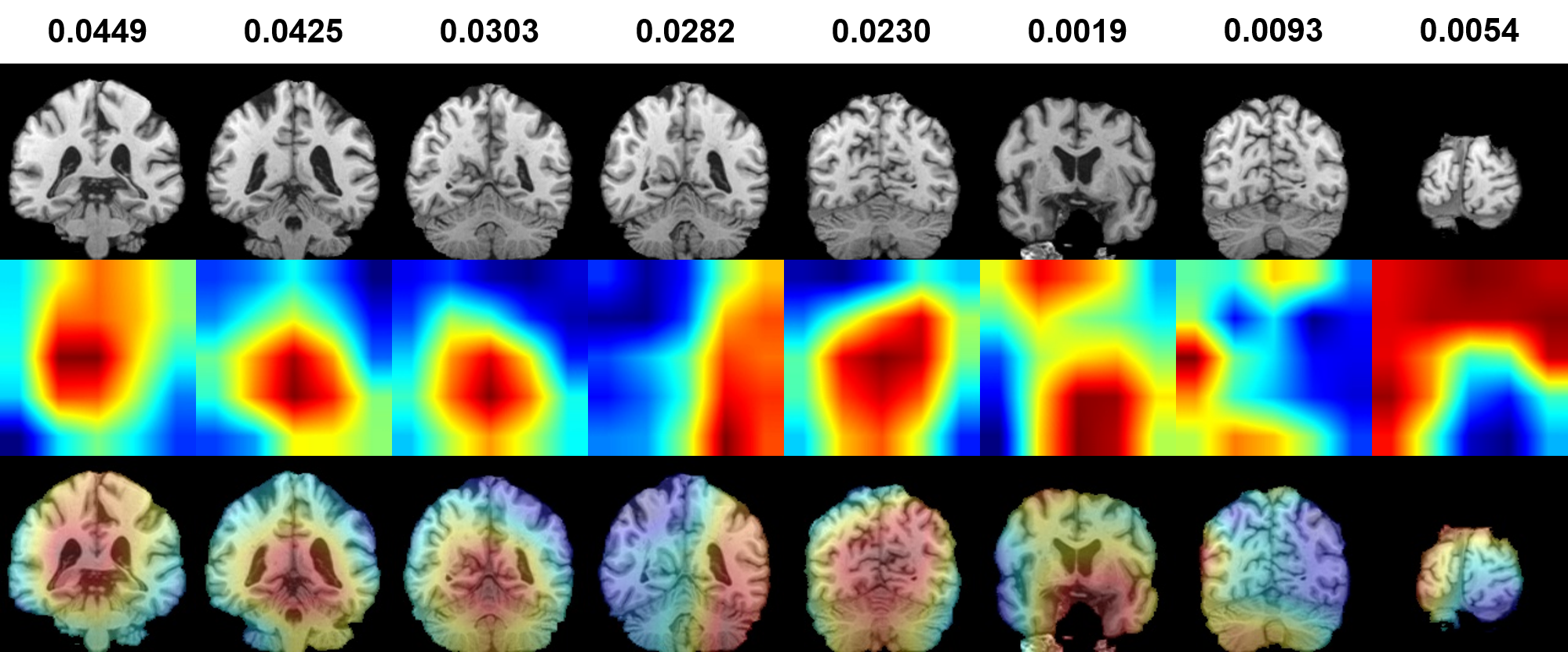}
    \caption{Qualitative examples of DGA-DMIL spatial attention maps with attention scores on sMRI samples from a single individual. The top row shows the attention value of each instance with the spatial maps normalized.}
    \label{fig:enter-label6}
\end{figure*}

\begin{figure}
    \centering
    \includegraphics[width=0.8\linewidth]{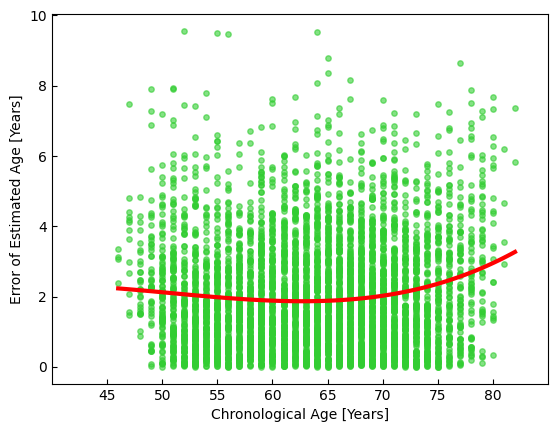}
    \caption{The distribution of the standard deviation $\sigma$ of the brain estimated ages computed by different individuals over the years. The red curve is the smooth of the average $\sigma$ on each year}
    \label{fig:enter-label7}
\end{figure}

\subsection{Qualitative Results with Multiple-instance Learning}
In Figure~\ref{fig:enter-label6}, we present qualitative results showcasing instance attention maps and attention scores. These results demonstrate the effectiveness of GDA-DMIL in identifying key slices associated with inferred areas using coarse maps. The heatmap illustrates the probability of obtaining the lowest MAE (best prediction) for the brain image. Interestingly, we observe low attention scores for slices containing noise or artifacts without any important areas, further highlighting the effectiveness of our method. Furthermore, the attention maps focus on crucial areas like ground-glass opacities and consolidations, which align well with clinical findings. The results indicate that the attention mechanism proposed in our approach remains relevant and holds great potential for clinical evaluation.

In Figure~\ref{fig:enter-label7} we visualize the distribution of the standard deviation ($\sigma$), which represents the uncertainty measure of the error in estimating brain age across individuals. A high $\sigma$ indicates significant variations in estimating brain age across different brain regions. Notably, the differences converge to a minimum across all age groups, indicating the robustness of our method across subjects of all ages.

\section{DISCUSSION} \label{sec:discussion}
In this study, we propose a novel framework, called DGA-DMIL, for brain age prediction using sMRI data. With the increasing interest in deep learning methods for computational analysis of sMRI data in brain age estimation, our approach stands out by accounting for the heterogeneity and redundant information in brain age estimation. To address this, we introduce a dual graph attention aggregator to assign different weights to different instances, thus contributing to the final age prediction results. It is capable of exploiting the rich instance feature maps for accurate age prediction while providing interpretability through the scores within the Multi-instance learning mechanism. Furthermore, we propose a disentanglement branch to enable independent learning of the age-related features and the age-independent structure representation. This disentanglement improves the accuracy of brain age estimation by preventing the interference of redundant information. Overall, our approach provides a comprehensive framework for brain age prediction by combining graph attention MIL pooling and decoupling representation learning.

In addition, our proposed method not only serves as a reliable tool for estimating brain age but also facilitates the interpretation of the evidence underlying age estimation. By using instance-level salient brain regions, we have successfully identified the contributions of brain regions to the final estimation results. To further validate the proposed approach, we provide qualitative evidence in the form of attention maps and MIL scores that highlight the specific regions of focus identified by the model. The attention maps obtained effectively highlight key areas in the majority of cases, with MIL scores corresponding to significant slices. These compelling results confirm the superiority of our proposed method in terms of performance compared to other models and provide visualized evidence for the estimated brain age. 

While our method has demonstrated good performance in estimating brain age on the UK Biobank and ADNI datasets, it is important to acknowledge the presence of certain limitations that should be addressed in future research. Firstly, our assumption that an individual's actual age corresponds to the biological age of their brain may not always hold. To improve the accuracy of age prediction, future work could explore the integration of age-related information from multiple organs, such as the heart and blood vessels. Secondly, the primary objective of this study is to develop an accurate estimation model for brain MRI. It is worth noting that, similar to other studies in the literature~\cite{he2021global, he2022deep, peng2021accurate}, patient MRIs are not included in this research. Thirdly, our proposed method focuses primarily on analyzing brain sMRI data and does not incorporate text-based information such as gender, race, or ethnicity, which could potentially affect the manifestation of brain age effects. In future studies, it would be valuable to investigate the inclusion of text-based variables to better understand the differential effects of brain age across different demographic groups. This could involve building separate models or accounting for the influence of gender, race, ethnicity, and other relevant factors. Additionally, imputation techniques could be used to further refine our method. By using imputation techniques, we can potentially increase the completeness and accuracy of the data, leading to improved brain age estimation performance.

\section{CONCLUSION} \label{sec:conclusion}
This paper proposes a Dual Graph Attention based Disentanglement MIL framework for brain age estimation based on sMRI. The proposed method efficiently captures and aggregates features of intra- and inter-instances within a bag. In addition, the proposed disentanglement branch is capable of extracting age and structural representations. The results show that our method achieves state-of-the-art performance and effectively identifies the most informative regions. The proposed method aims to make a valuable contribution to the existing literature on age estimation and provide additional insights into heterogeneous aging within an individual's brain.

\section*{ACKNOWLEDGMENTS}
This research has been conducted using the UK Biobank Resource under Application Number 94178.

\bibliographystyle{model2-names.bst}\biboptions{authoryear}
\bibliography{main}

\end{document}